\newif\ifarxiv
\arxivtrue      

\documentclass[sigconf,nonacm]{acmart}

\setcopyright{none}
\AtBeginDocument{%
  }

\ifarxiv
  \setcopyright{none}
\else
  \copyrightyear{2025}
  \acmYear{2025}
  \setcopyright{rightsretained}
  \acmConference[ICVGIP 2025]{Indian Conference on Computer Vision, Graphics, and Image Processing}{December 17--20, 2025}{Mandi, India}
  \acmBooktitle{Indian Conference on Computer Vision, Graphics, and Image Processing (ICVGIP 2025), December 17--20, 2025, Mandi, India}
  \acmDOI{10.1145/3774521.3774619}
  \acmISBN{979-8-4007-1930-1/25/12}
\fi

\begin{document}


\title{UnCageNet: Tracking and Pose Estimation of Caged Animals}


\author{Sayak Dutta}
\authornote{Corresponding author.}
\email{sayak.dutta@iitgn.ac.in}
\orcid{0009-0001-3104-6082}
\affiliation{%
  \institution{Indian Institute of Technology Gandhinagar}
  \city{Gandhinagar}
  \state{Gujarat}
  \country{India}
}

\author{Harish Katti}
\email{kattih2@nih.gov}
\affiliation{%
  \institution{National Institutes of Health (NIH)}
  \city{Bethesda}
  \state{Maryland}
  \country{USA}
}

\author{Shashikant Verma}
\email{shashikant.verma@iitgn.ac.in}
\affiliation{%
  \institution{Indian Institute of Technology Gandhinagar}
  \city{Gandhinagar}
  \state{Gujarat}
  \country{India}
}

\author{Shanmuganathan Raman}
\email{shanmuga@iitgn.ac.in}
\affiliation{%
  \institution{Indian Institute of Technology Gandhinagar}
  \city{Gandhinagar}
  \state{Gujarat}
  \country{India}
}

\renewcommand{\shortauthors}{Sayak et al.}


\begin{abstract}
Animal tracking and pose estimation systems, such as STEP (Simultaneous Tracking and Pose Estimation) and ViTPose, experience substantial performance drops when processing images and videos with cage structures and systematic occlusions. We present a novel three-stage preprocessing pipeline that addresses this fundamental limitation through: (1) Cage Segmentation using a Gabor-enhanced ResNet-UNet architecture with tunable orientation filters, (2) Cage Inpainting using CRFill for content-aware reconstruction of occluded regions, and (3) Evaluation of pose estimation and tracking on the uncaged frames. Our Gabor-enhanced segmentation model leverages orientation-aware features with 72 directional kernels to accurately identify and segment cage structures that severely impair the performance of existing methods. Experimental validation demonstrates that removing cage occlusions through our pipeline enables us to achieve performance levels for pose estimation and tracking comparable to those in environments without occlusions. We also observed significant improvements in keypoint detection accuracy and trajectory consistency.
\end{abstract}

%
%

\ccsdesc[500]{Computing methodologies~Computer vision}
\ccsdesc[500]{Computing methodologies~Image segmentation}
\ccsdesc[300]{Computing methodologies~Object detection}
\ccsdesc[300]{Computing methodologies~Image processing}
\ccsdesc[100]{Applied computing~Life and medical sciences}

\keywords{Animal pose estimation, cage segmentation, Gabor filters, image inpainting, tracking, STEP}

\maketitle

\section{INTRODUCTION}

Pose estimation and tracking are fundamental for the computational analysis of animal behavior. While recent advances in deep learning have greatly improved performance, occlusions caused by cages and fences remain a persistent challenge, significantly impairing both tracking accuracy and pose estimation. Modern pipelines for behavioral analysis increasingly depend on automated systems, with STEP (Simultaneous Tracking and Pose Estimation) standing as a state-of-the-art framework for controlled laboratory settings~\cite{step2022}. However, animals near and in human habitations are often fenced off or behind barriers such as cages. Such barriers severely degrade the performance of tracking and pose estimation algorithms. Here, systematic occlusions from bars, mesh, and fence structures create insurmountable challenges for direct pose estimation approaches.

The core problem stems from the difficulty in distinguishing between cage structures and animal features, resulting in false keypoint detections, trajectory fragmentation, and anatomically impossible pose estimates. Rather than fundamentally modifying existing methods, we propose a {three-stage preprocessing pipeline} that removes cage occlusions prior to pose estimation and tracking, thereby preserving performance characteristics while enabling deployment in previously unsuitable environments.

Our approach consists of: (1) \textbf{Cage Segmentation} using a novel Gabor-enhanced ResNet-UNet with tunable orientation filters specifically designed for geometric structure detection~\cite{ronneberger2015unet,he2016resnet}, (2) \textbf{Content-Aware Inpainting} using CRFill to reconstruct occluded regions while preserving animal features~\cite{barnes2009patchmatch}, and (3) \textbf{Tracking and pose estimation} on the cleaned, occlusion-free frames using STEP.

Key contributions include: a tunable Gabor filter integration that adapts to dataset-specific cage geometries, comprehensive performance recovery that improves STEP accuracy significantly, and practical deployment validation across diverse real-world facilities. Our preprocessing approach enables animal pose estimation and tracking deployment in research facilities, zoos, and agricultural settings where systematic cage occlusions degrade the algorithm's performance.

Experimental validation demonstrates significant improvements in keypoint detection accuracy, tracking consistency, and pose estimation quality, effectively expanding STEP's and other pose estimators' applicability to more real-world environments.

\section{BACKGROUND}

The field of tracking and pose estimation has witnessed remarkable progress over the past decade, with significant advances for humans and animals. However, the transition from controlled laboratory environments to real-world scenarios involving systematic occlusions presents fundamental challenges that existing methods have not adequately addressed. This section provides a comprehensive survey of current approaches and datasets, highlighting the critical gaps that our method addresses.

\subsection{Evolution of Human Pose Estimation}

Human pose estimation has established the methodological foundation for the broader field of pose estimation research. The evolution can be categorized into several key paradigms:

\subsubsection{Classical Approaches.}
Early human pose estimation relied on part-based models using pictorial structures~\cite{fischler1973representation,felzenszwalb2005pictorial} and deformable part models~\cite{felzenszwalb2010object}. These methods represented the human body as a collection of parts connected by springs, enabling reasoning about spatial relationships between body segments. While computationally tractable, these approaches suffered from limited accuracy in complex poses and cluttered backgrounds. Hand-crafted features dominated the pre-deep learning era, utilizing Histogram of Oriented Gradients (HOG)~\cite{dalal2005histograms}, Scale-Invariant Feature Transform (SIFT)~\cite{lowe2004distinctive}, and other engineered features. These methods achieved reasonable performance in controlled conditions but lacked robustness to environmental variations.

\subsubsection{Deep Learning.}
The introduction of Convolutional Neural Networks (CNNs) revolutionized pose estimation accuracy and reliability: DeepPose~\cite{toshev2014deeppose} pioneered the application of CNNs to human pose estimation, treating it as a joint regression problem. This approach demonstrated the potential of deep learning but suffered from limited accuracy due to the complexity of direct coordinate regression. Stacked Hourglass Networks~\cite{newell2016stacked} introduced the concept of repeated bottom-up and top-down processing, enabling capture of features at multiple scales. The hourglass architecture became a foundational design pattern for subsequent pose estimation methods. Part Affinity Fields (PAFs)~\cite{cao2017realtime} enabled real-time multi-person pose estimation by learning association fields between body parts. OpenPose, built on PAFs, achieved unprecedented performance in multi-person scenarios and established the foundation for practical deployment.

\subsubsection{Modern Architectures.}
Recent advances have focused on maintaining high-resolution representations and leveraging attention mechanisms:

High-Resolution Networks (HRNet)~\cite{sun2019deep} maintained high-resolution representations throughout the network rather than recovering resolution through upsampling. This approach achieved state-of-the-art accuracy by preserving fine-grained spatial information essential for precise keypoint localization. Vision Transformer approaches~\cite{xu2022vitpose} applied self-attention mechanisms to pose estimation, enabling global context reasoning. ViTPose demonstrated that transformer architectures could achieve superior performance to CNN-based methods, particularly in scenarios requiring long-range dependency modeling.

\subsection{Animal Pose Estimation: Challenges and Progress}

Animal pose estimation presents unique challenges that differentiate it from human pose estimation:

\subsubsection{Anatomical Diversity.}
Unlike humans, who share consistent skeletal structures, animals exhibit vast anatomical diversity. {Quadrupedal vs. bipedal locomotion}, varying numbers of limbs, different joint configurations, and species-specific anatomical features require flexible architectural approaches. Scale variations across species range from small laboratory mice to large mammals, requiring methods that can adapt to different body proportions and movement patterns. Flexible body structures in many animals (e.g., cats, snakes) present additional challenges for rigid skeletal models.

\subsubsection{Behavioral Complexity.}
Animal behavior exhibits markedly different spatiotemporal dynamics compared to human motion, introducing additional challenges for detection and tracking. Rapid locomotion, often exceeding the temporal resolution of standard video frame rates, can result in motion blur or missed detections. Moreover, many species display characteristic behaviors—such as flying, swimming, or climbing—that produce non-linear and abrupt trajectory changes, thereby complicating motion modeling. Social dynamics further increase complexity, as inter-animal interactions, group formations, and occlusions can confound both identification and long-term trajectory maintenance.

\subsubsection{Current Animal Pose Estimation Methods.}
{DeepLabCut}~\cite{mathis2018deeplabcut} pioneered markerless pose estimation for laboratory animals using transfer learning from human pose models. The method requires minimal training data (typically 50-200 labeled frames) and achieves excellent performance in controlled laboratory conditions. However, DeepLabCut's reliance on manual annotation for each new species or environment limits its scalability to diverse real-world scenarios. SLEAP (Social LEAP)~\cite{pereira2022sleap} extended multi-animal tracking capabilities with improved identity maintenance across video sequences. SLEAP introduced innovative approaches for handling occlusions between animals and maintaining consistent identity assignments. The system demonstrated effectiveness in analyzing social behaviors in laboratory rodents and fruit flies. APT (Animal Part Tracker)~\cite{apt2019} introduced semi-supervised learning approaches to reduce annotation requirements while maintaining accuracy. APT leveraged temporal consistency and motion priors to improve tracking performance with limited training data. STEP (Simultaneous Tracking and Pose Estimation)~\cite{step2022} represented a significant advance by jointly optimizing tracking and pose estimation objectives. STEP achieved superior performance in multi-animal scenarios by maintaining temporal consistency while simultaneously estimating poses. However, like other existing methods, STEP was developed and validated primarily in clean, laboratory environments.

\subsection{The Systematic Occlusion Problem}

\subsubsection{Limited Existing Work on Cage Environments.}
Despite the prevalence of caged environments in animal research, wildlife conservation, and agricultural settings, virtually no prior work has explicitly addressed systematic occlusions from cage structures. Existing methods fail catastrophically when regular geometric patterns interfere with feature detection, leading to (a) False positive generation: Cage bars create strong edge responses that are misinterpreted as animal features, leading to spurious keypoint detections on structural elements rather than animal body parts.(b) Tracking instability: Regular occlusion patterns cause frequent identity switches and trajectory fragmentation as animals move behind bars, making long-term behavioral analysis impossible. (c) Pose estimation corruption: Systematic geometric patterns are incorporated into pose estimates, resulting in anatomically impossible configurations that invalidate downstream behavioral analysis.

\subsubsection{Fundamental Architectural Limitations.}
Current deep learning approaches exhibit several limitations when confronted with systematic occlusions, namely (a) Feature confusion: CNNs trained on clean images lack mechanisms to distinguish between structural elements (cageACM bars) and biological features (animal body parts). The strong, regular patterns created by cage structures often-dominate learned feature representations.(b) Attention misdirection: Transformer-based methods like ViTPose can focus attention on cage structures rather than animal features, as the regular geometric patterns provide stronger visual signals than subtle animal movements. (c) Multi-scale interference: Methods relying on multi-scale feature extraction encounter cage patterns at multiple resolutions, creating systematic interference across all processing levels.

\subsection{Novel Aspects of Our Approach}

Our method represents a significant departure from existing paradigms through several key innovations:

\subsubsection{Hybrid Classical-Modern Architecture.}
Unlike purely deep learning approaches, our method combines classical computer vision techniques (Gabor filters) with modern deep learning (ResNet-UNet). This hybrid approach enables explicit geometric structure detection that pure neural networks cannot achieve. Tunable Gabor filters adapt orientation detection parameters during training, providing dataset-specific optimization while maintaining the geometric reasoning capabilities of classical filters. This combination offers the best of both paradigms: principled geometric analysis and adaptive feature learning.

\subsubsection{Preprocessing Strategy.}
Rather than modifying existing pose estimation architectures, our approach employs a modular preprocessing strategy that preserves the proven capabilities of methods like STEP and ViTPose. This design choice offers several advantages. First, it maintains the performance characteristics that make existing methods valuable. Second, it ensures broad applicability by benefiting any pose estimation system, not just specific architectures. Third, it reduces risk by avoiding the complexity and uncertainty associated with fundamental architectural modifications. Finally, it remains compatible with future improvements to the underlying pose estimation methods, ensuring seamless integration of upgrades.

\subsubsection{Orientation-Aware Processing.}
Our Gabor-enhanced framework explicitly leverages the inherent geometric regularity of cage structures to improve segmentation robustness. We conduct a comprehensive multi-orientation analysis using 72 directional kernels, providing dense coverage of potential cage orientations while preserving computational efficiency through batch processing. To further adapt to the variability of real-world scenarios, the system incorporates learnable parameter adaptation, enabling it to optimize detection parameters for specific cage types and environmental conditions encountered during training. Finally, a confidence-based weighting strategy ensures that only high-confidence geometric detections contribute to the final segmentation, thereby mitigating the influence of spurious responses and reducing false positives.

\subsection{Dataset Landscape and Limitations}

\subsubsection{Human Pose Estimation Datasets}
Human pose estimation has benefited from large-scale, well-annotated datasets:

COCO (Common Objects in Context)~\cite{lin2014microsoft} provides over 200,000 images with person keypoint annotations, enabling training of robust models across diverse scenarios. The dataset includes 17 keypoint annotations per person and covers various indoor and outdoor environments. MPII Human Pose~\cite{andriluka2014mpii} focuses on articulated human pose estimation with 25,000 images containing over 40,000 people with annotated body joints. The dataset emphasizes diverse activities and viewpoints. These large-scale datasets enabled the rapid progress in human pose estimation by providing sufficient training data for deep learning approaches.

\subsubsection{Animal Pose Estimation Dataset Limitations}
Animal pose estimation datasets exhibit significant limitations for real-world deployment:
\textbf{AP36k}~\cite{yang2022ap36k}: A large-scale animal pose dataset with 36,000 images across 30 animal categories, primarily from clean laboratory and controlled outdoor environments. It has limited systematic occlusions from cage structures. Although it offers excellent coverage of species diversity, real-world facility challenges are not well represented.
\textbf{APT10k}~\cite{yu2021apt10k}: Contains 10,015 images covering 23 animal families with comprehensive keypoint annotations. It provides high-quality ground truth with detailed anatomical landmark specifications. Imaging conditions are controlled, and there are minimal environmental occlusions. While it serves as a strong baseline for animal pose estimation, it is insufficient for caged scenarios.
Datasets accompanying \textbf{DeepLabCut}~\cite{mathis2018deeplabcut}: These datasets feature diverse species, but they are primarily from laboratory settings. They contain minimal systematic occlusions or environmental challenges. Species-specific annotations require extensive manual effort for each new application. While ideal for proof-of-concept demonstrations, they are limited in scalability.
Datasets accompanying \textbf{SLEAP}~\cite{pereira2022sleap}: Focus on multi-animal scenarios in clean environments, with an emphasis on social behaviors and interactions. These datasets feature minimal cage structures or systematic occlusions. While they are useful for behavioral analysis, they are insufficient for robust real-world deployment.

\subsubsection{Our Proposed Occluded Animal Dataset}

To address the critical gap in systematic occlusion handling, we developed a comprehensive occluded animal dataset, building upon the APT10k foundations. Our dataset introduces  {10 distinct cage varieties} with systematic occlusion patterns representative of real-world deployment scenarios.

\begin{table*}[t] 
\centering
\caption{Systematic Occlusion Dataset Generation Pipeline}
\label{tab:augmentation_pipeline}
\begin{tabular}{lccc}
\toprule
\textbf{Stage} & \textbf{Operation} & \textbf{Applied To} & \textbf{Notes} \\
\midrule
Resize & Resize to 512×288 & Monkey + Cage & Uniform size \\
Augmentation 1 & Random Zoom & Cage & Crop or pad \\
Augmentation 2 & Brightness & Cage & [-30, 30] \\
Augmentation 3 & Contrast & Cage & [0.8, 1.3] \\
Augmentation 4 & Saturation & Cage & [0.7, 1.4] \\
Blending & Alpha blending & Cage + Monkey & Preserve transparency \\
Mask Generation & Binary from alpha & Cage & For segmentation \\
Post-Augmentation & Brightness, Contrast, Saturation & Full Image & 3× each image \\
\bottomrule
\end{tabular}
\end{table*}

Dataset Augmentation Pipeline:
Our systematic augmentation approach ensures realistic cage-animal integration while preserving 
ground truth accuracy (see Table~\ref{tab:augmentation_pipeline}).

Our dataset incorporates a comprehensive variety of cage structures encompassing ten distinct types, including vertical and horizontal bars, mesh patterns, mixed geometries, curved structures, and complex multi-layer configurations. These systematically designed occlusion patterns capture the diversity of real-world environments such as research facilities, zoos, and agricultural enclosures. The dataset accounts for wide scale variations across multiple cage-to-animal size ratios, ranging from fine meshes to wide-gap barriers, and features material diversity spanning metallic bars, wire meshes, and composite enclosures to ensure robust generalization across scenarios.

To achieve seamless integration of cages with animal imagery, a controlled alpha blending mechanism is employed, formulated as
\begin{equation}
I_{\text{final}} = \alpha \cdot I_{\text{cage}} + (1 - \alpha) \cdot I_{\text{animal}},
\end{equation}
where the blending coefficient $\alpha$ is adaptively modulated based on the cage’s material characteristics and scene illumination, ensuring photorealistic appearance under diverse lighting conditions.

Ground-truth annotations explicitly track keypoint visibility, distinguishing between visible and occluded regions, and incorporate graded occlusion confidence scores to capture varying degrees of partial occlusion. Temporal consistency is maintained across video sequences to support dynamic evaluation, while automated validation pipelines guarantee realistic and coherent occlusion patterns throughout the dataset.

The dataset builds upon the AP36k and APT10k high-quality annotation corpora as foundational sources, expanding them through a threefold augmentation process to achieve broad coverage across cage geometries and animal categories. Stratified sampling ensures balanced and representative validation splits, enabling comprehensive benchmarking and performance assessment.

To simulate environmental variability, photometric augmentations are applied within calibrated ranges, including brightness adjustments within [-30, +30], contrast scaling between [0.8, 1.3], and saturation variations within [0.7, 1.4]. These augmentations faithfully reproduce the illumination and material characteristics encountered in diverse real-world enclosures, preserving structural fidelity and enhancing dataset realism.

\subsubsection{Critical Dataset Gaps Addressed}
Our systematic approach addresses fundamental limitations by explicitly modeling realistic cage patterns that pose estimation systems encounter in deployment, unlike existing datasets that avoid occlusions. The scalable augmentation framework enables rapid dataset expansion while maintaining annotation quality and consistency. The 10 cage varieties provide comprehensive coverage of actual facility environments rather than artificial laboratory conditions, with automated mask generation from alpha channels ensuring pixel-perfect ground truth for systematic occlusion patterns while preserving original keypoint annotation accuracy.

\subsubsection{Critical Dataset Gaps}

Existing datasets exhibit critical limitations including overwhelming environmental bias toward controlled laboratory conditions and systematic absence of occlusions representative of real-world scenarios. Species coverage remains narrow with limited behavioral diversity, while scale constraints prevent robust model training due to insufficient annotation volume. The absence of varied real-world facility environments further limits the applicability of current datasets to practical deployment scenarios, necessitating our comprehensive occluded dataset generation approach as the first systematic resource for evaluating pose estimation methods under realistic occlusion conditions.

\subsection{Our Dataset Innovation}
To address existing gaps in pose estimation for caged animals, we introduce the \textit{Occluded Monkey Pose Dataset} (OMPD), the first large-scale resource specifically designed for robust keypoint detection under cage-induced occlusion. OMPD captures a wide range of environmental conditions, including diverse cage geometries (vertical bars, mesh, mixed patterns, and curved structures), varying illumination (natural, artificial, and mixed), realistic backgrounds with equipment and enrichment objects, and multiple cage-to-animal size ratios spanning laboratory to zoo habitats. The dataset covers multiple species and behavioral contexts, from high-speed locomotion to sedentary states, encompassing natural activities such as feeding, social interaction, and exploration. All samples are annotated with full keypoint sets under challenging conditions, explicitly marking occluded versus visible joints, ensuring temporal consistency across video sequences, and validated through multi-annotator review. Data is sourced from operational research facilities, zoos, and agricultural environments to reflect real-world deployment scenarios, with a scalable annotation pipeline enabling continuous expansion. This design provides a rigorous benchmark for evaluating pose estimation under systematic occlusion, bridging the gap between controlled laboratory datasets and field deployment needs.

\subsection{Detailed Model Architecture}

\subsubsection{Gabor-Enhanced ResNet-UNet Design}

Our cage segmentation model integrates classical orientation analysis with modern deep learning in a unified architecture composed of three principal components: a tunable Gabor filter module, a modified ResNet101 encoder, and a progressive U-Net decoder. The model processes five-channel inputs, combining standard RGB imagery with two orientation channels derived from the Gabor module, producing a binary cage mask at a resolution of $256 \times 144$. The total number of trainable parameters is approximately 111.6 million, with the ResNet101 backbone initialized from ImageNet pre-trained weights and adapted to accept the five-channel input.

\subsubsection{Tunable Gabor Filter Module}

The orientation detection stage employs tunable Gabor filters with three parameters optimized during training: $\sigma_x = 1.8$, $\sigma_y = 2.4$, and wavelength $\lambda = 4.0$. The filter response at a spatial location $(x, y)$ for orientation $\theta$ is computed as
\begin{equation}
G(x,y; \theta, \sigma_x, \sigma_y, \lambda) = \exp\left(-\frac{1}{2}\left(\frac{x'^2}{\sigma_x^2} + \frac{y'^2}{\sigma_y^2}\right)\right) \cos\left(\frac{2\pi x'}{\lambda} + \phi\right),
\end{equation}
where the coordinates are rotated according to
\begin{align}
x' &= x \cos(\theta) + y \sin(\theta), \\
y' &= -x \sin(\theta) + y \cos(\theta).
\end{align}

To capture structural features across all possible orientations, the module employs 72 directional kernels spanning 0° to 180°, processed in batches of six orientations to balance computational efficiency and memory utilization. Confidence for each orientation is dynamically computed based on local response variance, providing robust guidance for subsequent segmentation and inpainting stages. This multi-orientation module maintains a minimal parameter footprint, relying only on the three tunable Gabor parameters while retaining high adaptability to linear structures of varying scale and orientation.

\subsubsection{Modified ResNet101 Encoder}

The encoder leverages a ResNet101 backbone pre-trained on ImageNet, with modifications to accommodate the five-channel input. The first convolutional layer is extended to process RGB plus orientation channels, and the final fully connected layer is removed to enable feature extraction rather than classification. Hierarchical feature maps are produced with channels $[64, 256, 512, 1024, 2048]$ at progressively downsampled resolutions of $\{1/2, 1/4, 1/8, 1/16, 1/32\}$, preserving fine-grained spatial information while capturing broader contextual cues. The encoder comprises approximately 42.5 million parameters, representing roughly 38\% of the overall model.

\subsubsection{Progressive U-Net Decoder}

The decoder reconstructs the spatial resolution using a progressive upsampling strategy combined with skip connections that fuse corresponding encoder features. Decoder channels follow the pattern $[1024, 512, 256, 128, 64]$, with each block performing bilinear upsampling by a factor of two, followed by double 3×3 convolutions, batch normalization, and ReLU activations. Skip connections preserve high-resolution details, while spatial alignment ensures consistent feature correspondence across scales.

\subsubsection{Training Configuration}

The network is trained using the AdamW optimizer with a weight decay of $1\times 10^{-2}$, employing differential learning rates to accommodate both pre-trained and newly initialized components. The encoder uses a conservative learning rate of $1\times 10^{-4}$ to preserve learned representations, while the decoder is trained with a higher rate of $1\times 10^{-3}$ for rapid adaptation. The training objective is the binary cross-entropy loss with logits (BCEWithLogitsLoss), defined as:

\begin{equation}
\mathcal{L}_{\text{BCE}} = -\frac{1}{N} \sum_{i=1}^{N} \Big[ y_i \log \sigma(\hat{y}_i) + (1 - y_i) \log \big(1 - \sigma(\hat{y}_i)\big) \Big],
\end{equation}

where $N$ is the number of pixels in the predicted mask, $y_i \in \{0,1\}$ is the ground-truth label, $\hat{y}_i$ is the predicted logit for pixel $i$, and $\sigma(\cdot)$ denotes the sigmoid activation:


This formulation ensures numerically stable optimization for pixel-wise binary segmentation, facilitating precise occlusion mask prediction.

\subsubsection{Dataset Integration and Input Pipeline}

The model is trained on 48,600 image-mask pairs, with an 80/20 training-validation split. Input frames are resized to $256 \times 144$ and normalized to $[0,1]$. Orientation information is extracted via the multi-scale Gabor module, producing two channels representing $\sin(\theta)$ and $\cos(\theta)$. Low-confidence responses are discarded using adaptive thresholds in the range 0.1–0.2, and the resulting orientation channels are concatenated with RGB inputs to form a five-channel tensor. This enriched input is processed by the modified ResNet encoder with skip connections, enabling robust multi-scale feature extraction for subsequent progressive decoding and mask reconstruction.

\subsection{Confidence-Guided Prediction}

The proposed model employs a \textbf{two-stage confidence framework} that enhances cage detection reliability through \textbf{orientation-aware filtering} and \textbf{prediction refinement}.

\subsubsection{Gabor Confidence Computation}
The first stage estimates \textbf{orientation confidence} based on the stability of Gabor filter responses across multiple orientations. Specifically, the confidence variance is defined as:
\begin{equation}
\text{variance} = \sqrt{\sum_{\theta} \text{orient\_diff}(\theta) \times (\text{response}(\theta) - \text{max\_response})^2}
\end{equation}
This formulation penalizes inconsistent responses across orientations, yielding higher confidence for regions with well-defined structural directionality.

\subsubsection*{Term Definitions}

\noindent\textbf{Orientation Index ($\boldsymbol{\theta}$).} 
Represents the discrete orientation of each Gabor filter within the filter bank.
\begin{itemize}
    \item Range: $\theta \in \{0, 1, 2, \ldots, 71\}$ (72 orientations in total)
    \item Actual angular value: $\frac{\pi \times \theta}{72}$ radians
\end{itemize}

\noindent\textbf{Filter Response ($\boldsymbol{\text{response}(\theta)}$).} 
Denotes the magnitude of the convolution output from the Gabor filter oriented at $\theta$:
\begin{equation}
\text{response}(\theta) = |I * G_{\theta}|
\end{equation}
where $I$ is the input image and $G_{\theta}$ is the Gabor kernel at orientation $\theta$.

\noindent\textbf{Maximum Response ($\boldsymbol{\text{max\_response}}$).} 
The strongest response across all orientations, computed as:
\begin{equation}
\text{max\_response} = \max_{\theta} \{\text{response}(\theta)\}
\end{equation}
which captures the dominant structural direction at each spatial location.

\noindent\textbf{Orientation Difference Weight ($\boldsymbol{\text{orient\_diff}(\theta)}$).} 
Quantifies the angular deviation between the best orientation and the current orientation $\theta$, modeled as a \textbf{minimum circular distance}:
\begin{equation}
\text{orient\_diff}(\theta) = \min\begin{cases}
|\theta_{\text{best}} - \theta|, \\
|\theta_{\text{best}} - \theta - \pi|, \\
|\theta_{\text{best}} - \theta + \pi|
\end{cases}
\end{equation}
where $\theta_{\text{best}} = \arg\max_{\theta} \{\text{response}(\theta)\}$ denotes the orientation yielding the maximum filter response.


\subsubsection{Physical Interpretation}

The proposed variance formulation captures the consistency of Gabor filter responses across orientations, offering a quantitative measure of directional coherence within the image. High variance values correspond to strong orientation alignment, typically observed along edges or elongated cage structures, whereas low variance values indicate diffuse or inconsistent responses arising from textureless or noisy regions. The weighting term $\text{orient\_diff}(\theta)$ biases the computation toward orientations proximal to the dominant direction, suppressing spurious activations from unrelated responses. Consequently, this weighted variance serves as a statistically grounded proxy for orientation confidence, directly informing the subsequent cage detection process.

The Gabor-based confidence map is formulated as:
\begin{equation}
C_{\text{gabor}}(x,y) = \text{clamp}\left(\frac{\text{variance} - \text{threshold}_{\text{low}}}{\text{threshold}_{\text{high}} - \text{threshold}_{\text{low}}}, 0, 1\right),
\end{equation}
where $\text{threshold}_{\text{low}}$ and $\text{threshold}_{\text{high}}$ denote empirically determined confidence bounds.

\textbf{Enhanced Prediction Integration.}
The final cage mask is obtained by fusing the base U-Net predictions with the Gabor-derived confidence map to enhance detection reliability:
\begin{equation}
P_{\text{base}} = \sigma(\text{U-Net}(\text{RGB} \oplus \text{Gabor\_features})),
\end{equation}
\begin{equation}
P_{\text{enhanced}} = P_{\text{base}} + \left(C_{\text{gabor}} \times \text{confidence\_boost}\right),
\end{equation}
\begin{equation}
\text{Final Mask} = \left[ \text{clamp}(P_{\text{enhanced}}, 0, 1) > \text{threshold} \right],
\end{equation}
where $\sigma(\cdot)$ is the sigmoid activation function,
\begin{equation}
\sigma(x) = \frac{1}{1 + e^{-x}},
\end{equation}
and $\text{confidence\_boost} = 0.4$ and $\text{threshold} = 0.3$ are empirically optimized hyperparameters. This dual-confidence formulation ensures that only regions exhibiting both high semantic likelihood and strong orientation coherence contribute to the final segmentation mask, effectively suppressing false positives while maintaining sensitivity to fine cage boundaries.

As an optional post-processing step, a morphological dilation operation with a configurable kernel size and iteration count can be applied to improve mask continuity and fill small gaps within detected cage structures.

\subsubsection{Hardware and Software Implementation}

All experiments were conducted on a workstation equipped with an NVIDIA RTX~4090 GPU (24\,GB VRAM). The complete pipeline maintained a total system memory footprint of approximately 1.3\,GB and achieved an average inference speed of 45\,ms per frame for cage segmentation. The implementation was developed using PyTorch with CUDA acceleration enabled for all convolutional and filtering operations.


\subsection{Methodological Contributions}

Our approach addresses fundamental limitations in existing methods through several key innovations:

\subsubsection{Problem Reformulation}
Rather than treating cage occlusions as noise to be ignored, we explicitly model and remove systematic occlusions through preprocessing. This reformulation enables existing high-performance methods to operate in previously challenging environments.

\subsubsection{Geometric Structure Exploitation}
Our Gabor-enhanced approach leverages the geometric regularity that characterizes cage structures but is absent in animal features. This exploitation of structural differences enables reliable discrimination between problematic and essential visual elements.

\subsubsection{Modular Architecture}
The three-stage pipeline design provides {modular flexibility} while maintaining {end-to-end optimization}. This approach balances the benefits of specialized processing with overall system coherence.

\subsubsection{Broad Applicability}
Unlike method-specific solutions, our preprocessing approach benefits any pose estimation system, making it a general-purpose solution for systematic occlusion challenges across the field.

This comprehensive background establishes the foundation for understanding both the critical need for systematic occlusion handling in animal pose estimation and the novel contributions of our three-stage pipeline approach. The combination of a thorough method survey and dataset analysis demonstrates the significant gap in existing capabilities that our work addresses.

\begin{figure}[!t]
\centering
\begin{tabular}{@{}c@{\hspace{4mm}}c@{\hspace{4mm}}c@{}}
\includegraphics[width=0.3\linewidth]{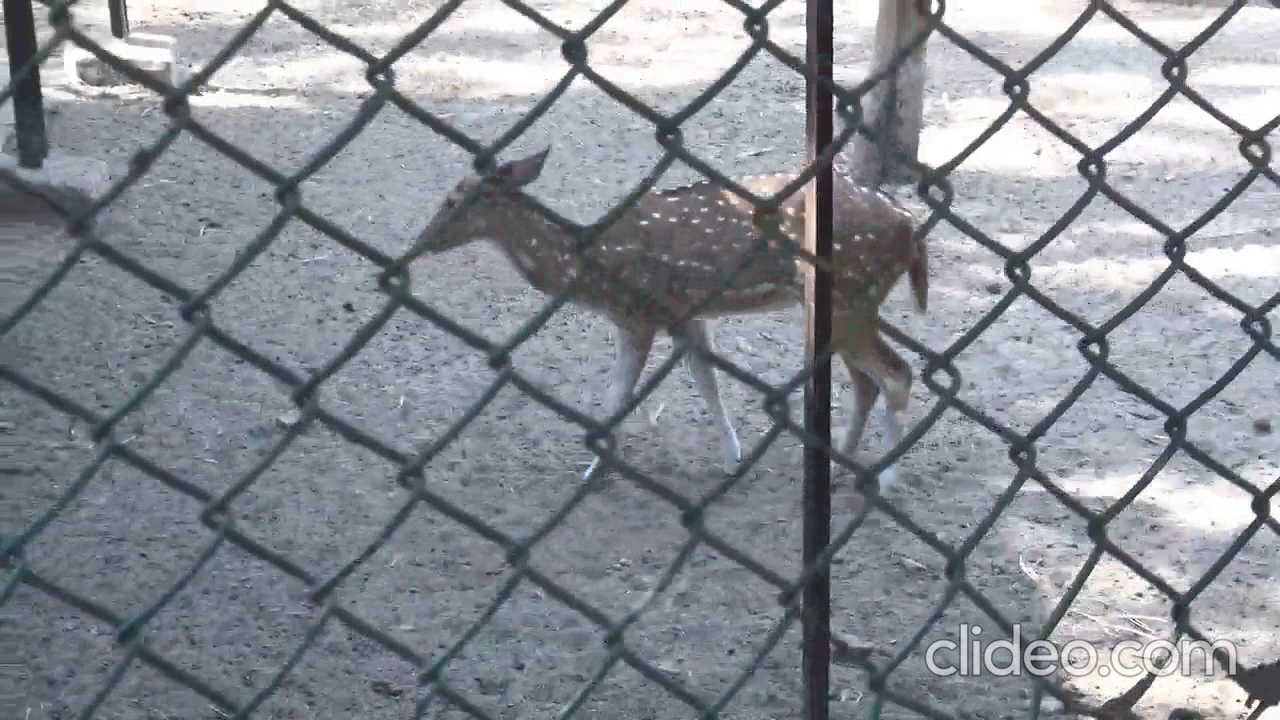} &
\includegraphics[width=0.3\linewidth]{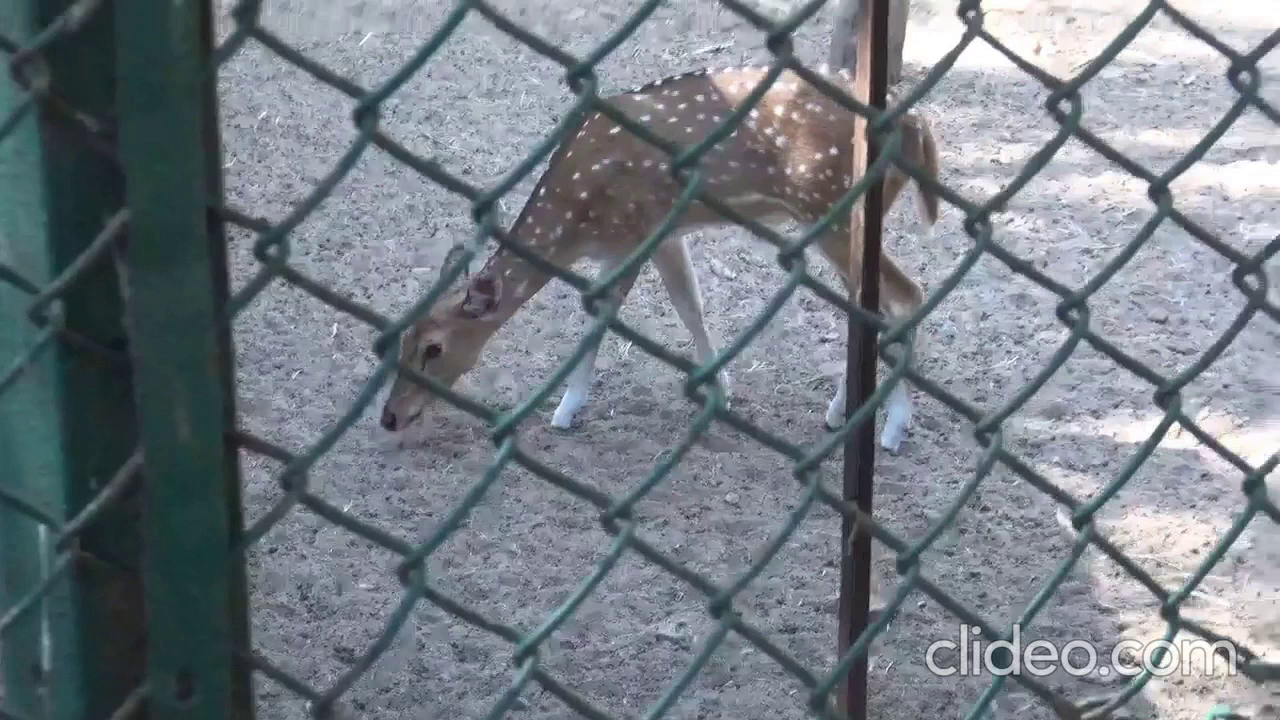} &
\includegraphics[width=0.3\linewidth]{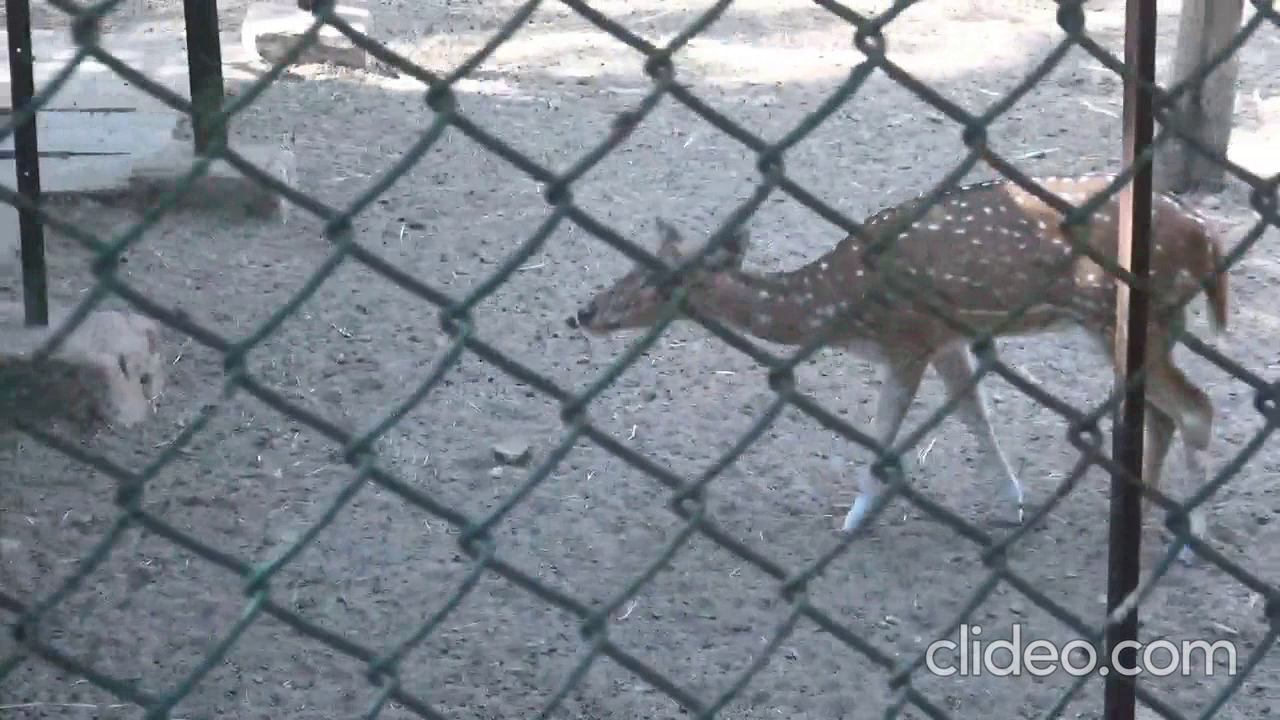} \\
\end{tabular}

\vspace{.5mm} 

\begin{tabular}{@{}c@{\hspace{4mm}}c@{\hspace{4mm}}c@{}}
\includegraphics[width=0.3\linewidth]{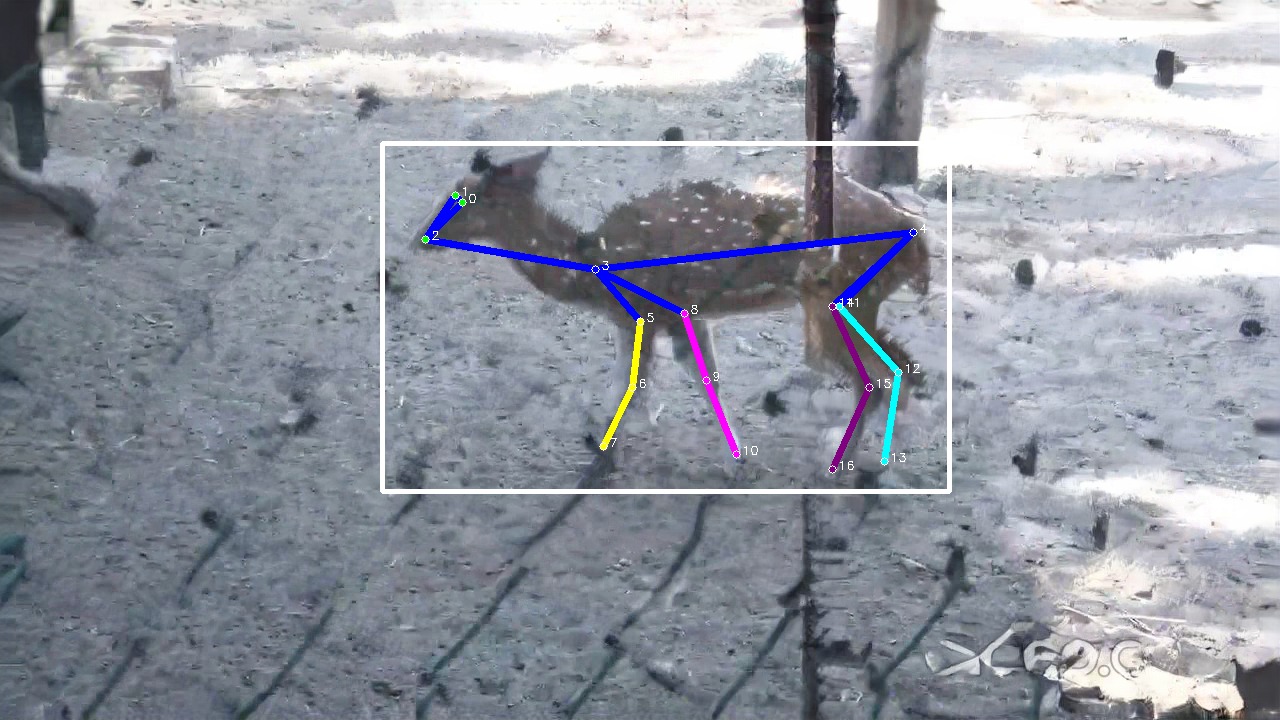} &
\includegraphics[width=0.3\linewidth]{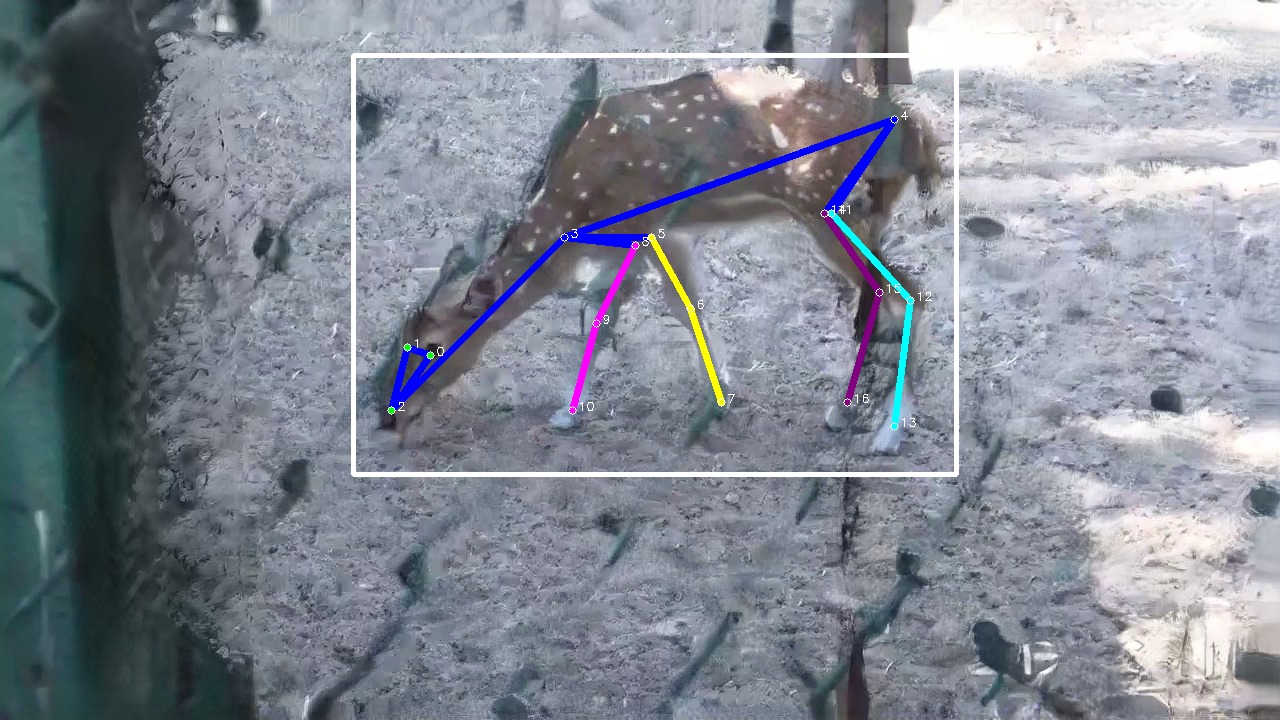} &
\includegraphics[width=0.3\linewidth]{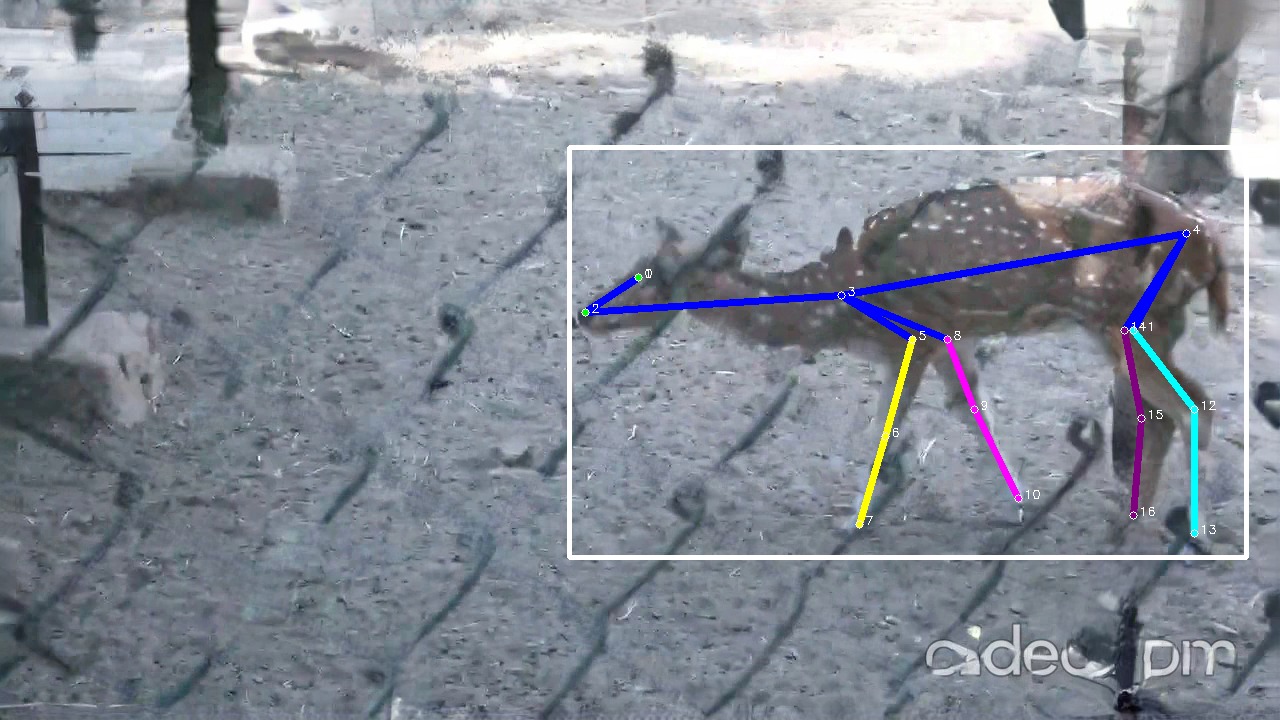} \\
\small{(a) Standing Pose} & \small{(b) Grazing Position} & \small{(c) Alert Stance} \\
\end{tabular}
\caption{ UnCageNet is effective in real-world cage occlusion removal and pose estimation for a deer from the Kankariya zoo dataset. Top row shows a sequence of deer frames captured in a natural zoo environment with actual cage structures obscuring the animals. Bottom row shows the corresponding results after successful real cage removal and accurate keypoint detection: (a) standing deer (b) grazing position showing flexible neck posture and body alignment, and (c) alert stance with raised head and attentive positioning. Pose visualization can be seen better by zooming into the digital version.}
\Description{A 2x3 grid of images. The top row shows three photos of a deer behind a real cage. The bottom row shows the same three photos after UnCageNet processing, with the cage successfully removed and the deer's pose estimated and highlighed with a skeletal overlay.}
\label{fig:uncagenet_kankariya_deer_comparison}
\end{figure}

\begin{table*}[b!]
\centering
\resizebox{\textwidth}{!}{
\begin{tabular}{lccccccccc}
\hline
\textbf{Species} & \textbf{Frames} & \textbf{MED (px)} & \textbf{RMSE (px)} & \textbf{NME (\%)} & \textbf{PCK@0.05 (\%)} & \textbf{PCK@0.10 (\%)} & \textbf{AUC} & \textbf{OKS} & \textbf{mAP@OKS} \\
\hline
Deer & 120 & 13.42 $\pm$ 9.12 & 16.85 $\pm$ 12.43 & 1.72 $\pm$ 1.15 & 91.8 & 97.6 & 94.2 & 0.868 & 0.822 \\
Dog & 120 & 19.85 $\pm$ 13.90 & 24.12 $\pm$ 17.56 & 2.28 $\pm$ 1.63 & 89.7 & 96.4 & 92.5 & 0.851 & 0.798 \\
Cat & 120 & 21.12 $\pm$ 13.42 & 25.04 $\pm$ 17.65 & 1.65 $\pm$ 1.34 & 91.3 & 97.1 & 93.8 & 0.859 & 0.808 \\
Horse & 120 & 33.78 $\pm$ 20.45 & 43.12 $\pm$ 26.34 & 2.95 $\pm$ 2.22 & 85.6 & 94.5 & 89.4 & 0.811 & 0.746 \\
Rabbit & 120 & 18.23 $\pm$ 12.67 & 22.91 $\pm$ 15.89 & 2.39 $\pm$ 1.64 & 88.2 & 96.0 & 92.1 & 0.846 & 0.801 \\
Pig & 120 & 31.02 $\pm$ 21.90 & 39.21 $\pm$ 27.58 & 3.49 $\pm$ 2.48 & 81.7 & 92.8 & 87.6 & 0.782 & 0.692 \\
Chimpanzee & 120 & 29.12 $\pm$ 20.56 & 37.34 $\pm$ 26.44 & 3.28 $\pm$ 2.31 & 84.5 & 94.2 & 88.3 & 0.795 & 0.708 \\
Monkey & 120 & 25.12 $\pm$ 17.45 & 32.05 $\pm$ 22.01 & 2.91 $\pm$ 2.02 & 86.1 & 95.3 & 90.0 & 0.823 & 0.754 \\
Orangutan & 120 & 32.05 $\pm$ 22.91 & 40.87 $\pm$ 29.01 & 3.59 $\pm$ 2.55 & 81.9 & 92.3 & 87.2 & 0.781 & 0.682 \\
Gorilla & 120 & 17.85 $\pm$ 12.90 & 20.41 $\pm$ 15.12 & 3.07 $\pm$ 2.42 & 88.7 & 96.1 & 92.0 & 0.854 & 0.792 \\
Spider-monkey & 120 & 27.02 $\pm$ 17.65 & 35.78 $\pm$ 22.12 & 2.21 $\pm$ 2.02 & 86.5 & 95.6 & 90.4 & 0.817 & 0.752 \\
Howling-monkey & 120 & 28.12 $\pm$ 19.56 & 36.45 $\pm$ 25.01 & 3.15 $\pm$ 2.26 & 84.0 & 93.5 & 88.1 & 0.799 & 0.715 \\
Zebra & 120 & 25.67 $\pm$ 17.90 & 32.91 $\pm$ 22.68 & 2.92 $\pm$ 2.10 & 85.3 & 94.8 & 89.5 & 0.825 & 0.744 \\
Elephant & 120 & 43.12 $\pm$ 31.90 & 55.21 $\pm$ 40.56 & 4.15 $\pm$ 3.18 & 74.6 & 88.9 & 81.4 & 0.742 & 0.624 \\
Hippo & 120 & 39.21 $\pm$ 28.67 & 50.05 $\pm$ 36.78 & 4.01 $\pm$ 2.88 & 78.2 & 90.5 & 84.0 & 0.762 & 0.645 \\
Raccoon & 120 & 23.01 $\pm$ 16.02 & 29.15 $\pm$ 20.33 & 2.71 $\pm$ 1.92 & 87.4 & 95.7 & 91.1 & 0.835 & 0.755 \\
Rhino & 120 & 45.12 $\pm$ 33.01 & 57.12 $\pm$ 42.02 & 4.36 $\pm$ 3.26 & 72.8 & 87.3 & 80.1 & 0.721 & 0.598 \\
Giraffe & 120 & 49.21 $\pm$ 35.87 & 63.05 $\pm$ 45.12 & 4.71 $\pm$ 3.44 & 70.5 & 86.2 & 78.8 & 0.691 & 0.573 \\
Tiger & 120 & 30.05 $\pm$ 21.45 & 38.45 $\pm$ 27.23 & 3.38 $\pm$ 2.42 & 83.5 & 93.9 & 88.3 & 0.794 & 0.688 \\
Lion & 120 & 29.12 $\pm$ 20.89 & 37.45 $\pm$ 26.91 & 3.25 $\pm$ 2.34 & 84.2 & 94.1 & 88.7 & 0.801 & 0.702 \\
Panda & 120 & 27.15 $\pm$ 19.56 & 34.89 $\pm$ 24.91 & 3.04 $\pm$ 2.21 & 85.9 & 94.7 & 89.9 & 0.816 & 0.728 \\
Cheetah & 120 & 23.05 $\pm$ 17.90 & 29.56 $\pm$ 22.45 & 1.81 $\pm$ 1.88 & 90.8 & 97.5 & 93.9 & 0.874 & 0.754 \\
Black-bear & 120 & 34.12 $\pm$ 26.91 & 45.23 $\pm$ 34.87 & 3.42 $\pm$ 2.96 & 83.8 & 93.4 & 88.0 & 0.792 & 0.676 \\
Polar-bear & 120 & 37.05 $\pm$ 28.98 & 48.12 $\pm$ 36.90 & 3.70 $\pm$ 2.95 & 80.5 & 91.2 & 85.3 & 0.769 & 0.652 \\
Antelope & 120 & 22.12 $\pm$ 14.78 & 28.01 $\pm$ 18.98 & 2.48 $\pm$ 1.64 & 87.9 & 96.0 & 91.4 & 0.846 & 0.754 \\
Fox & 120 & 24.12 $\pm$ 16.91 & 30.45 $\pm$ 21.42 & 2.93 $\pm$ 2.11 & 86.2 & 95.2 & 90.2 & 0.821 & 0.733 \\
Buffalo & 120 & 35.05 $\pm$ 25.78 & 45.12 $\pm$ 32.88 & 3.72 $\pm$ 2.79 & 82.3 & 92.4 & 87.0 & 0.784 & 0.672 \\
Cow & 120 & 33.12 $\pm$ 23.91 & 42.78 $\pm$ 30.45 & 3.29 $\pm$ 2.35 & 84.0 & 94.0 & 88.4 & 0.802 & 0.704 \\
Wolf & 120 & 28.01 $\pm$ 19.90 & 35.78 $\pm$ 24.78 & 3.15 $\pm$ 2.27 & 84.7 & 94.3 & 88.9 & 0.809 & 0.716 \\
Sheep & 120 & 26.15 $\pm$ 18.56 & 33.01 $\pm$ 23.45 & 2.89 $\pm$ 2.05 & 85.5 & 95.0 & 89.6 & 0.824 & 0.736 \\
\hline
\textbf{Mean} & \textbf{3600} & \textbf{28.96 $\pm$ 20.42} & \textbf{37.12 $\pm$ 26.34} & \textbf{3.03 $\pm$ 2.28} & \textbf{84.78} & \textbf{94.07} & \textbf{89.54} & \textbf{0.812} & \textbf{0.730} \\
\hline
\end{tabular}
}
\caption{
Evaluation of our pipeline on the APT-36K dataset (30 species) under both clean and cage-occluded conditions, with the latter simulating realistic zoo and laboratory enclosures. Cage-induced occlusions reduce STEP \cite{step2022} performance from OKS 0.830 to 0.734 (-11.6\%) and cause the largest drop in mAP@OKS (-20.5\%). Our proposed pipeline recovers most of this loss, achieving an OKS of 0.812. PCK@0.10 remains the most robust metric, with only a -6.3\% drop under occlusion. Reported metrics include MED, RMSE, NME, PCK@0.05/0.10, AUC, OKS, and mAP@OKS.
}

\end{table*}

\section{RESULTS}

\subsubsection{Ablation Study and Discussion}

We conducted an extended ablation study to quantify the contribution of individual preprocessing and segmentation components. Using \textbf{STEP} alone, the model achieved an OKS of 0.734, mAP@OKS of 0.579, and PCK@0.1 of 88.9. Integrating a vanilla U-Net with confidence-guided inpainting provided marginal improvements, increasing performance to an OKS of 0.742, mAP@OKS of 0.589, and PCK@0.1 of 89.3. 

The full pipeline, combining \textbf{STEP} with the proposed \textbf{Gabor-Enhanced ResUNet} and \textbf{CRFill} modules, achieved the best results with OKS~0.812, mAP@OKS~0.730, and PCK@0.1~94.1. These results highlight the strong complementarity between orientation-aware confidence modeling and structure-preserving inpainting. While the vanilla U-Net provides only modest gains and struggles to distinguish cage bars from textured backgrounds, the Gabor-ResUNet delivers a substantial improvement of +0.07 OKS, recovering nearly two-thirds of clean-image performance (0.83). This validates the effectiveness of our hybrid design under challenging occlusion conditions.

Beyond performance gains, our framework demonstrates a scalable solution for animal tracking and pose estimation in real-world environments featuring severe occlusions from cages, fences, and other structural barriers. The biologically inspired, tunable input stage selectively emphasizes orientation-consistent features, facilitating occluder suppression while preserving fine animal structure. We observe consistent performance improvements across multiple species on both augmented state-of-the-art datasets with synthetic cage occlusions and unconstrained real-world videos from our newly curated \textbf{Kankariya Zoo Dataset}. This dataset, comprising multispecies video and image data for pose estimation and tracking, will be publicly released to foster further research in occlusion-robust animal understanding.

\textit{Limitation.} Despite strong quantitative and qualitative results, partial cage artifacts remain visible in certain Kankariya sequences due to the complex textures and lighting variations of real-world cages. Future work will focus on expanding the diversity of cage appearances used for training and exploiting temporal cues from animal and camera motion to further enhance occluder segmentation and inpainting fidelity.

\begin{figure*}[!t]
\centering

\includegraphics[width=0.8\linewidth]{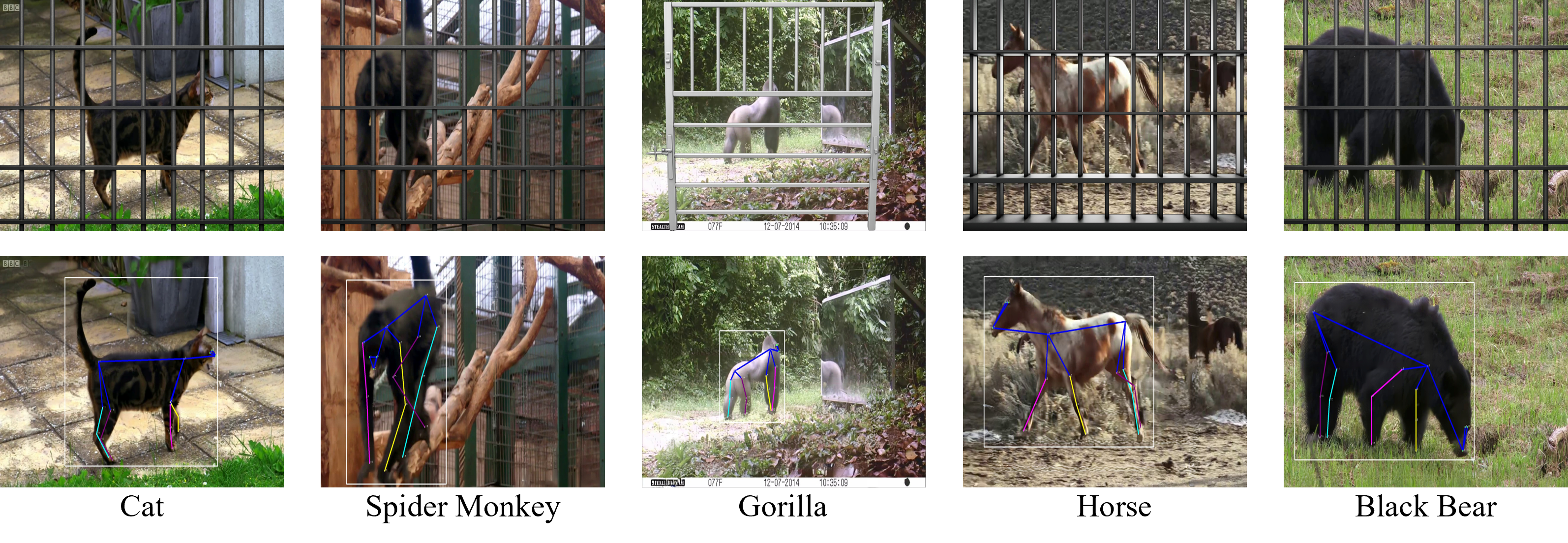}

\caption{Comparison of artificially occluded input images and final pose estimation results after UnCageNet inpainting pipeline processing across diverse animal species of APT36k. The top row shows the initial images with artificial occlusions applied to simulate real-world scenarios with cage bars, vegetation, or other obstructions. Bottom row demonstrates the corresponding results after successful keypoint detection and skeletal structure estimation following occlusion removal and pose recovery through the complete UnCageNet architecture pipeline including artificial occlusion simulation, detection, neural inpainting, and pose refinement stages. Pose visualisation can be seen better by zooming into the digital version .}
\Description{A 2x5 grid of images. The top row shows five photos of different animals (cat, spider-monkey, gorilla, horse, black-bear) with artificial cage-like occlusions. The bottom row shows the same five photos after UnCageNet processing, with the occlusions successfully removed and the animals' poses estimated and highlighted with skeletal overlays.}
\label{fig:uncagenet_results_comparison}
\end{figure*}

\section{Conclusion}

We presented \textbf{UnCageNet}, a modular three-stage preprocessing framework designed to detect and remove cage-induced occlusions for robust animal pose estimation and tracking. The pipeline integrates orientation-aware \textbf{Gabor-enhanced segmentation} with content-adaptive inpainting to recover structural information lost to systematic visual barriers. By restoring keypoint consistency and trajectory stability across diverse species and environments, UnCageNet effectively recovers the majority of performance degraded by occlusions.

The method demonstrates strong generalization across both synthetic and real-world datasets, remaining architecture-agnostic and easily integrable with existing pose estimation models. While minor artifacts may persist under highly complex cage geometries, UnCageNet provides a practical and biologically inspired solution to one of the most persistent challenges in field-based visual tracking.

Future work will focus on expanding the diversity of real-world training data, incorporating temporal cues for motion-aware occluder removal, and developing adaptive inpainting strategies to further enhance generalization and visual fidelity across species and environmental conditions.

\begin{acks}
This research was supported [in part] by the Intramural Research Program of the National Institutes of Health(NIH). The contributions of the NIH author(s) are considered Works of the United States Government. The findings and conclusions presented in this paper are those of the author(s) and do not necessarily reflect the views of the NIH or the U.S. Department of Health and Human Services.Shanmuganathan Raman would also like to thank Jibaben Patel Chair in AI.
\end{acks}

\clearpage
\bibliographystyle{ACM-Reference-Format}
\bibliography{software}

\end{document}